\documentclass[tinyml]{acmart}
\usepackage{xspace, enumitem, makecell}%

\AtBeginDocument{%
  \providecommand\BibTeX{{%
    \normalfont B\kern-0.5em{\scshape i\kern-0.25em b}\kern-0.8em\TeX}}}

\setcopyright{rightsretained}
\copyrightyear{2025} %
\acmYear{2025} %

\begin{document}

\title{SplitQuant: Layer Splitting for Low-Bit Neural Network Quantization}

\author{Jaewoo Song}
\email{jsongab@connect.ust.hk}
\orcid{0000-0003-0082-7967}
\affiliation{%
  \institution{Department of Computer Science and Engineering \protect\\ The Hong Kong University of Science and Technology}
  \state{Hong Kong}
  \country{China}
}
\affiliation{%
  \institution{Nota Inc.}
  \city{Seoul}
  \country{Republic of Korea}
}

\author{Fangzhen Lin}
\email{flin@cse.ust.hk}
\affiliation{%
  \institution{Department of Computer Science and Engineering \protect\\ The Hong Kong University of Science and Technology}
  \state{Hong Kong}
  \country{China}
}

\newcommand{\splitquant}{SplitQuant\xspace}
\begin{abstract}
Quantization for deep neural networks (DNNs) is the process of mapping the parameter values of DNNs from original data types to other data types of lower precision to reduce model sizes and make inference faster.
Quantization often maps different original values to a single quantized value because the range of the original values is larger than the range of the quantized values.
This leads to the degradation of the accuracy of the quantized DNNs.
Outliers are a main cause of the degradation of quantization resolution because they enlarge the range of original values.
To solve the problem, the percentile method is often used to clip outliers.
However, clipping the outliers has another problem of removing the important and strong signals in the DNNs.
This paper proposes \splitquant to keep the outliers and improve the quantization resolution at the same time.
\splitquant narrows down the range of the original values and mitigates the effect of outliers by splitting each quantizable layer into three mathematically equivalent layers and applies different scaling factors.
Especially, weights and biases are clustered into lower, middle and upper clusters for optimized split.
By preprocessing DNNs with \splitquant, quantization algorithms can achieve better results.
\splitquant was applied on two BERT-Tiny models and improved the accuracy of INT2 quantization by 3.3\%p and 2.1\%p, achieving accuracies comparable to those of the original FP32 models.
\end{abstract}

\maketitle

\section{Introduction}
The number of parameters of deep neural network (DNN) models has been increasing.
Models with more parameters require more memory and storage spaces, and takes more inference time due to increased amount of calculation.
If the 32-bit or 64-bit floating-point parameter data types (FP32 or FP64) are converted to 2-bit, 4-bit or 8-bit integers (INT2, INT4 or INT8), both space and time can be saved.
For example, INT2 takes only 6.25\% and 3.125\% of space compared to FP32 and FP64.
Also, integer calculation is 3 to 5 times faster than floating-point calculation~\cite{tvm_2019}.

The process of converting floating-point parameters to integers is called quantization.
Mathematically, quantization is the mapping process from the domain of floating-point to the codomain of integer.
For instance, FP32 to INT8 quantization maps 32-bit floating-point numbers in the range $[-3.4 \cdot 10^{38}, 3.4 \cdot 10^{38}]$ to 8-bit integers in the range $[-128, 127]$.
The process often suffers from irreversible errors because 32 bits has to be mapped to 8 bits.
Such errors make quantized DNN models less accurate.
Therefore, how to reduce such errors is a main research area of quantization.

Outliers are one of the main causes of quantization errors.
Outliers are input values far away from the mean when the input distribution is given.
Simply speaking, outliers are values much bigger or smaller than other values.
The performance of the quantization function is heavily affected by outliers because outliers worsen the resolution of the quantization function.
Conversely, quantization can be performed well if there are no outliers.

\sloppy For example, $[-1000.0, -500.0, 0.0, 500.0, 1000.0]$ can be quantized as $[-10, -5, 0, 5, 10]$ to fit in the target range $[-10, 10]$.
This is a good quantization because different original values are mapped to different target values.
However, if the input values are $[-1000.0, -500.0, 0.0, 500.0, 1.0 \cdot 10^{30}]$, the quantization result will be $[-10, -10, -10, -10, 10]$ because the outlier $1.0 \cdot 10^{30}$ is so big that the differences between other numbers become negligible, and thus lowers quantization resolution.
The quantization result becomes bad because four different original values -1000.0, -500.0, 0.0 and 500.0 are all mapped to the same value, -10.

Percentile clipping is the method to prevent the problem caused by outliers.
It ignores input values exceeding certain percentile (often 99\% is used in practice) when deciding the clipping range $[\beta, \alpha]$.
Although it is a de facto approach to deal with outliers in practice, there is a problem.
Outliers in DNN models convey strong signals.
Outliers in weights and biases state that certain features in the neural layer should be very sensitive to certain inputs.
And outliers in activation values mean that such features were indeed highly activated by an input.
Therefore, ignoring outliers causes losing important signals and may cause a negative effect on the accuracy of DNN models.

So here is a dilemma.
If outliers are kept, the quantization result may not be good.
On the other hand, if they are ignored, important signals may be lost.

This paper proposes \splitquant to solve the dilemma.
\splitquant solves the problem by splitting a layer to three mathematically equivalent layers to narrow down the ranges of original values.
Weights and biases are split via k-means clustering for optimized split.
They can be clustered because their values are known in advance.
Then the original layer is replaced by three layers created from the clusters.
Activations cannot be clustered because activation values are not known at quantization time.
Even if calibration data are given, it is not guaranteed that the calibration data represent all possible inputs.
Nonetheless, \splitquant still can achieve better quantization resolution by splitting the original layer to three layers because the ranges of the split layers will be narrower than the original range.

One of the important aspects of \splitquant is that it is intended to complement, not compete with, other quantization methods.
By reshaping DNN models into more quantization-friendly structures, \splitquant allows other quantization algorithms to achieve better results.

\splitquant was applied on two fine-tuned BERT-Tiny models~\cite{devlin2018bert}~\cite{turc2019well} from Hugging Face~\cite{wolf2019huggingface}.
The first model was fine-tuned on DAIR.AI's emotion recognition dataset~\cite{saravia2018carer}, and the second models was fine-tuned on UC Irvine SMS Spam Collection dataset for spam detection~\cite{almeida2011contributions}.
\splitquant enhanced the accuracies of INT2 quantizations for the models by 3.3\%p and 2.1\%p, respectively.
This led to accuracy increases from 86.5\% to 89.8\% and from 96.2\% to 98.3\% for each of the INT2 quantized models.
Given that the original FP32 models had accuracies of 90.2\% and 98.4\%, \splitquant improved the INT2 quantization accuracies close to those of the original FP32 models.

In summary, the dilemma of quantization caused by outliers and how \splitquant solves the problem are as below.
\begin{itemize}[leftmargin=*]
\item Outliers convey important signals, but at the same time worsens the quantization performance by lowering the quantization resolution.
\item \splitquant splits each quantizable layer to three mathematically equivalent layers to keep outliers and achieve finer quantization resolution at the same time.
\item \splitquant uses k-means clustering to optimize the split for weights and biases.
\item \splitquant can be used together with other quantization algorithms to help them achieve better quantization results.
\item Tested on two fine-tuned BERT-Tiny language models, \splitquant achieved 3.3\%p and 2.1\%p improvement in accuracies for INT2 quantization.
\end{itemize}

\section{Related Works}
In regard to the related works, it should be noted that \splitquant is not to compete with other quantization algorithms.
Rather, \splitquant reshapes original DNN models to be more quantization-friendly so that other quantization algorithms can achieve better results.

Net2Net~\cite{chen2015net2net} introduced the concept of function-preserving transformations to enhance DNN training.
It consists of Net2WiderNet and Net2DeeperNet for transforming a small ``teacher'' network to a wider or deeper ``student'' network.
The student network is functionally equivalent with the teacher network and has more neurons.
So, the knowledge of the teacher network is transferred to the student network, and the student network can be trained further by exploring the enlarged parameter space.
\splitquant differs from Net2Net because \splitquant is not for training but for quantization.
Also \splitquant does not always increase the number of neurons.
For example, \splitquant splits the activation layer into three layers but keeps the total number of neurons same.
On the other hand, Net2Net always increase the number of neurons to enlarge the parameter space.

Cell division~\cite{park2019cell} uses Net2Net approach to reduce the bit-width of weight parameters of convolutional neural networks (CNNs).
While it specifically targets the weight of CNNs only, \splitquant is applicable for the weight, bias and activation of any kind of neural networks.

OCS~\cite{zhao2019improving} also applied Net2Net to reduce the magnitude of outliers and improve quantization.
It does so by duplicating a neuron and then halving its output or outgoing weights.
Halving the outgoing weights requires modification of the input of the neuron.
\splitquant differs from OCS because \splitquant is not based on Net2Net.
Also, while OCS focuses on outliers only, \splitquant approach can still improve the quantization resolution even when there are no outliers.

VS-Quant~\cite{dai2021vs} applies separate scaling factors for vectors of elements of a tensor.
It shares similar idea with \splitquant, which is to use separate scaling factors instead of one scaling factor for a tensor.
VS-Quant focused much on modifying the hardware to support per-vector scaling.
In contrast, \splitquant can readily be used on conventional hardware and does not require any additional hardware support because \splitquant achieves the idea of separate scaling factors by reshaping the input model.

Again, it is important to emphasize that \splitquant is not to compete but to help other quantization algorithms, including the related works, by creating mathematically equivalent and more quantization-friendly DNN models.
It is totally possible, and in fact encouraged, to use \splitquant together with other quantization approaches to get better quantization results.

\section{Quantization and Outliers}
It is crucial to know about the quantization process to understand why outliers increase errors.
Let's consider the quantization process of mapping floating-point values in range $[\beta, \alpha]$ to integer values of bit-width $b$ in range $[-2^{b-1}, 2^{b-1}-1]$ as an example.
It can be mathematically expressed as
\begin{eqnarray}
Q(x)&=&\mathrm{INT}\left( Sx \right) + Z\\
S&=&\frac{2^{b}-1}{\alpha - \beta}\\
Z&=&-2^{b-1} - \mathrm{INT}\left( S \beta \right)
\end{eqnarray}
where $x$ is the FP32 input value, $S$ is a scaling factor, $\mathrm{INT}()$ is a rounding function and $Z$ is an offset called zero-point.

Scaling factor $S = \frac{2^{b}-1}{\alpha - \beta}$ scales the original values in range $[\beta, \alpha]$ to the quantization range $[-2^{b-1}, 2^{b-1}-1]$.
It is very important to note that the magnitude of the scaling factor determines the resolution of quantization.
If the scaling factor is too small, in other words $\alpha - \beta$ is too big for the given bit-width $b$, the resolution of the quantization function worsens and many different original values are mapped to a same quantization value.

Zero-point $Z$ is an offset which corresponds to the target value to which 0 in the original domain will be mapped.
Under the condition $\alpha = - \beta$, $Z$ becomes 0 and the quantization function is simplified as $Q(x) = \mathrm{INT}(Sx)$.
This is called symmetric quantization, while the condition $\alpha \neq \beta$ leads to asymmetric quantization.

The quantized values can be restored back to the original values by the process called dequantization as below.
However, dequantization often suffers from errors because of the rounding involved in the $\mathrm{INT}()$ function.
\begin{eqnarray}
\hat{x}&=&\frac{Q(x) - Z}{S}\\
&=&\frac{\mathrm{INT}(Sx) + Z - Z}{S}\\
&=&\frac{\mathrm{INT}(Sx)}{S}
\end{eqnarray}

\section{\splitquant}
As noted in the previous section, the magnitude of the scaling factor $S = \frac{2^{b}-1}{\alpha - \beta}$ is positively related to the quantization resolution, and ultimately to the performance of quantization.
\splitquant improves quantization resolution by increasing the scaling factor.
Since the bit-width $b$ is fixed, \splitquant increases the scaling factor by decreasing $\alpha - \beta$.

\begin{figure}
    \includegraphics[width=0.45\textwidth]{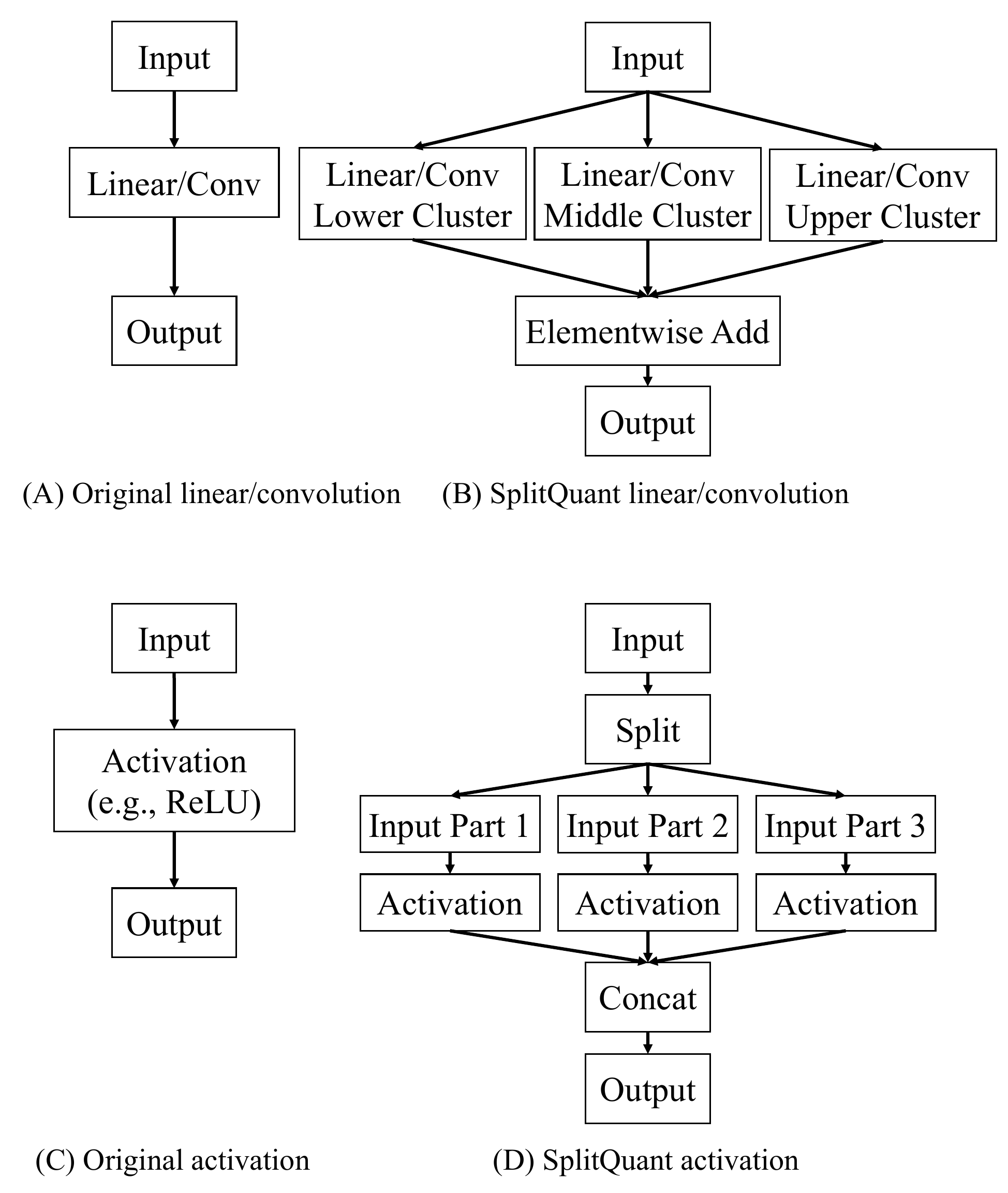}
    \caption{\splitquant splits each quantizable layer to three layers and combine them to improve quantization resolution while preserving functionality. (A) Original linear or convolution layer. (B) \splitquant splits the original linear or convolution layer by clustering the weights and biases. The results from the split layers are added elementwise so that the output will be preserved. (C) Original activation layer. (D) \splitquant splits the original activation and concatenate the results. Clustering is not possible for activation layers because the activation values can only be known in runtime. So, the original activation layer is divided into three activation layers, each with one-third of the original length. Then the results are concatenated to get the output of length.}
    \label{fig:how_to_split}
\end{figure}

\splitquant splits linear, convolution and activation layers and combine them while preserving the functionality of DNNs.
Since the distance between the maximum and minimum values in each layer (i.e., $\alpha - \beta$) is smaller than in the original layer, the quantization resolution is improved.
How \splitquant splits layers is graphically represented in Figure~\ref{fig:how_to_split}, and the details of how \splitquant optimizes the split is explained in the following subsections.

\subsection{\splitquant for weights and biases}
\begin{figure}
    \includegraphics[width=0.5\textwidth]{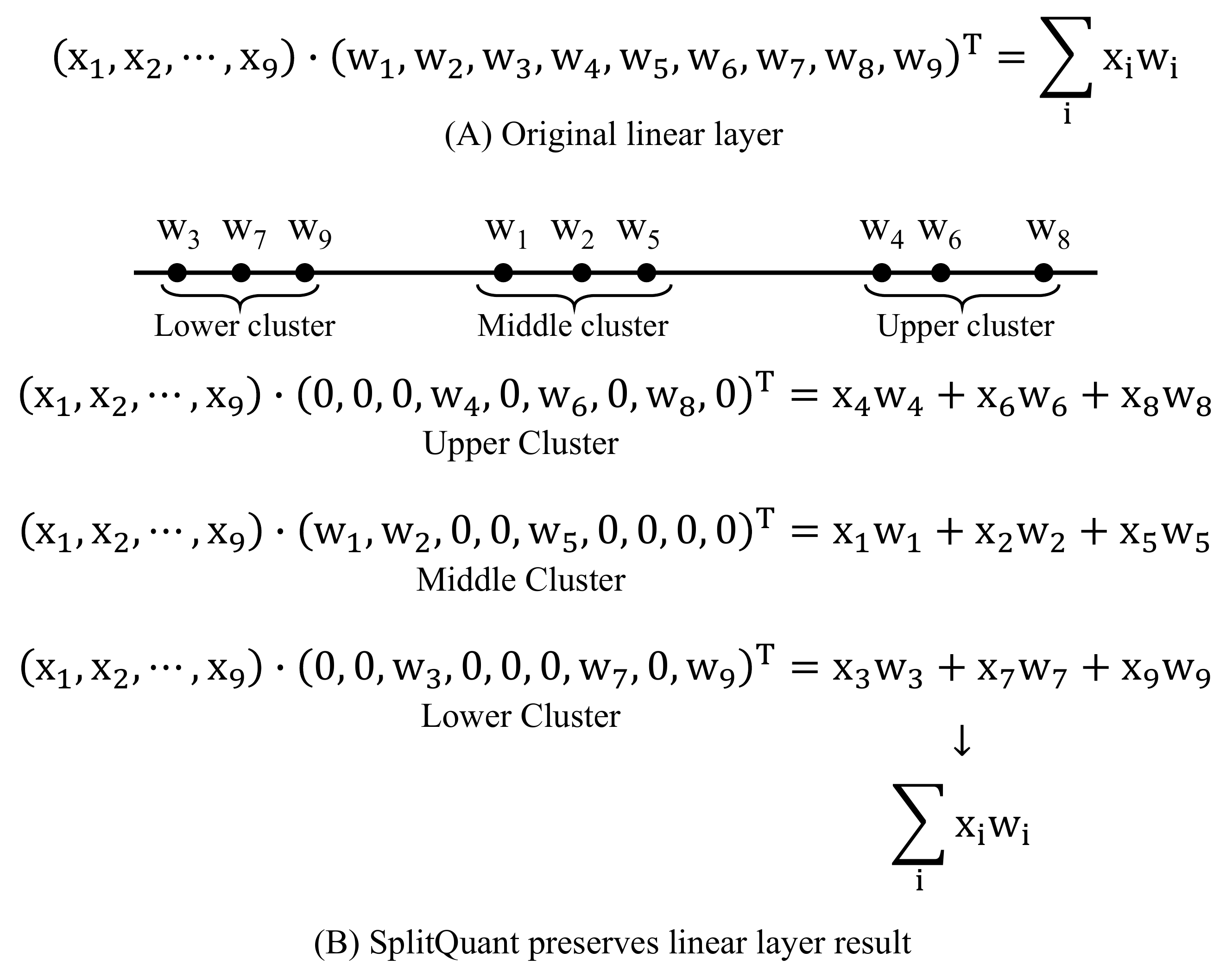}
    \caption{(A) Original linear layer. (B) \splitquant runs k-means clustering on the weights (and biases if exist) to cluster them into lower, middle and upper clusters. Then three new linear layers are created from the clusters. The original linear layer and the three split layers are mathematically equivalent. The split layers have higher quantization resolutions because their ranges are smaller than the range of the original layer.}
    \label{fig:split_linear}
\end{figure}

\begin{figure}
    \includegraphics[width=0.5\textwidth]{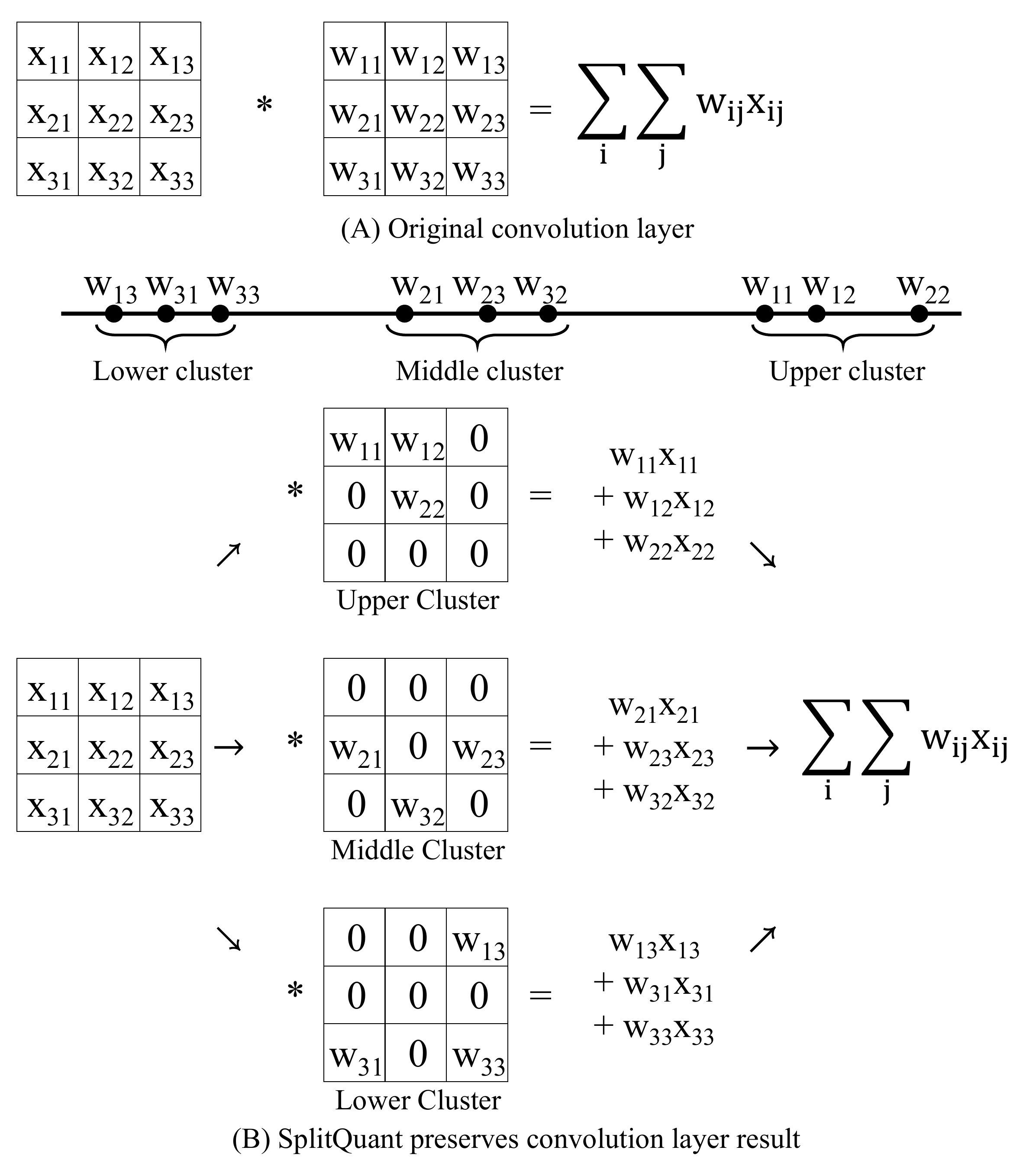}
    \caption{(A) Original convolution layer. (B) \splitquant runs k-means clustering on the weights (and biases if exist) to cluster them into lower, middle and upper clusters. Then three new linear layers are created from the clusters. The original convolution layer and the three split layers are mathematically equivalent. The split layers have higher quantization resolutions because their ranges are smaller than the range of the original layer.}
    \label{fig:split_conv}
\end{figure}

\splitquant runs k-means clustering on weights and biases.
With $k = 3$, weight and bias parameters are clustered into lower, middle and upper clusters.
Initial cluster centroids are selected by the greedy k-means++ algorithm~\cite{grunau2023nearly}.
Then three new layers are created from the clustered parameters.
For example, lower layer consists of the lower clusters of weight and bias parameters.
Then the original layer is replaced by the newly created layers.
The shapes of weights and biases in each new layer are maintained by injecting 0 where needed.
Figure~\ref{fig:split_linear} and Figure~\ref{fig:split_conv} show this idea for linear and convolution layers.

It is better to fold batch normalization layers into preceding linear and convolution layers before applying \splitquant.
It is because batch normalization folding reduces number of layers and hence reduces the error from quantization while preserving the functionality.
Also, it should be noted that PyTorch~\cite{paszke2019pytorch} internally represents the gamma parameters of normalization layers as weights.
Since they are semantically not weights, they should not be clustered.

\subsection{\splitquant for activation layers}
Activation layers cannot be clustered because the activation values can only be known in runtime.
Even if calibration data are given, the calibration data does not represent all real-world inputs.
Therefore, the original activation layer with length $n$ is split into three activation layers with length $n/3$.
Then the results are concatenated to get the output of length $n$.
Still, splitting activation layers can improve quantization because the ranges of activation values are narrowed down.

Let $\alpha$ and $\beta$ be the maximum and minimum activation values in the original layer.
If the original layer is split into three new layers and $\alpha$ and $\beta$ are put into different layers, then the scaling factors of all three split layers will increase and hence quantization resolution will be improved.
Even if $\alpha$ and $\beta$ are put into a same layer, two other layers' quantization resolution will still be improved.

Some quantization tools such as Quanto~\cite{huggingface_quanto} from Hugging Face support only weight quantization by default.
For such quantization methods, activation layers should not be split to avoid unnecessary split and concatenation operations.

\section{Results}
\begin{table*}[!ht]
\setlength\tabcolsep{2.5pt}
\begin{tabular}{c|c|ccc|ccc|ccc}
\toprule
BERT-Tiny & FP32 & \multicolumn{3}{c|}{INT2 Quantization} & \multicolumn{3}{c|}{INT4 Quantization} & \multicolumn{3}{c}{INT8 Quantization}\\
\hline
Dataset & Original & Baseline & \splitquant & Diff. & Baseline & \splitquant & Diff. & Baseline & \splitquant & Diff.\\
\midrule
\makecell{DAIR.AI\\Emotion\\Recognition} & 90.2\% & 86.5\% & 89.8\% & \underline{\textbf{+3.3\%p}} & 90.0\% & 90.2\% & +0.2\%p & 90.2\% & 90.3\% & +0.1\%p\\
\midrule
\makecell{UC Irvine\\SMS Spam} & 98.4\% & 96.2\% & 98.3\% & \underline{\textbf{+2.1\%p}} & 98.3\% & 98.4\% & +0.1\%p & 98.4\% & 98.4\% & 0.0\%p\\
\bottomrule
\end{tabular}
\caption{BERT-Tiny models fine-tuned on DIAR.AI's emotion recognition and UC Irvine's SMS spam detection datasets were quantized into INT2, INT4 and INT8 with and without \splitquant. The effect of \splitquant increased as the quantization bits decreased. This is because low-bit quantizations have reduced quantization resolution, making them more susceptible to outliers. For INT2 quantization, \splitquant achieved improvements of 3.3\%p and 2.1\%p in accuracies. It should be noted that the accuracies of INT2 quantization with \splitquant is close to the original accuracy of the FP32 models.}
\label{tab:result}
\end{table*}

\splitquant was applied on two fine-tuned BERT-Tiny models~\cite{gokuls_bert}~\cite{mrm8488_bert} downloaded from Hugging Face~\cite{wolf2019huggingface}.
The models were selected because BERT~\cite{devlin2018bert}~\cite{turc2019well} effectively represents the transformer architecture.
Also, their relatively small sizes make them suitable for tinyML and Edge AI.

DAIR.AI's emotion recognition dataset~\cite{saravia2018carer} and UC Irvine's SMS Spam Collection dataset for spam detection~\cite{almeida2011contributions} were used to fine-tune and test the models.
DAIR.AI's emotion recognition dataset~\cite{saravia2018carer} consists of train, validation and test datasets.
The test set of 2000 samples was used to figure out the effect of \splitquant on quantization.

UC Irvine's SMS Spam Collection dataset consists of 5574 samples and is not divided into subsets.
Therefore, the entire dataset used for fine-tuning the model was also utilized to compare the accuracies of the quantized models with and without \splitquant.
Since the goal of the experiment is to measure the improvement provided by \splitquant, the test setting is valid because the quantized models with and without \splitquant were compared in the same environment.

The effect of \splitquant was most dramatic for INT2 quantization.
For INT2 quantization, \splitquant improved the accuracy of the emotion recognition by 3.3\%p, increasing it from 86.5\% to 89.8\%.
The accuracy of the spam recognition was also improved by 2.1\%p, increasing from 96.2\% to 98.3\%.
The improved accuracies were very close to the original FP32 accuracies which were 90.2\% and 98.4\%.
\splitquant also improved the results for INT4 and INT8 quantizations, albeit less dramatic than for INT2.
The whole experiment result is shown in Table~\ref{tab:result}.

\section{Discussion and Future Work}
The effect of \splitquant was most significant for low-bit quantizations.
As can be seen in Table~\ref{tab:result}, \splitquant made most improvement for INT2 quantization.
The improvement of \splitquant for INT2 quantization was about 20 to 30 times more than the improvements for INT4 and INT8 improvements.
It is because low-bit quantizations has low quantization resolution, and hence are more susceptible to outliers.
And \splitquant successfully helped INT2 quantizations.

\splitquant increases the size of quantized models because there are three times more linear and convolution layers.
For FP32 to INT2 quantization as an example, the INT2 quantization model size will be 6.25\% of the original, while \splitquant can increase the INT2 quantization model size up to 18.75\% of the original.
Since newly added parameters by \splitquant are all 0s, the model size, memory usage and inference speed may be optimized if \splitquant is used together with sparse DNN inference engines such as SparseDNN~\cite{wang2021sparsednn}.

Finally, it will be an interesting research topic to apply \splitquant beyond the scope of TinyML and Edge AI such as large language models (LLMs).

\section{Conclusion}
\splitquant proves to be a highly effective method for improving the accuracy of low-bit quantizations, such as INT2 quantization, which are especially vulnerable to outliers due to their low quantization resolution.
By splitting each quantizable layer into three mathematically equivalent layers, \splitquant successfully keeps the important signals conveyed by outliers while simultaneously enhancing the quantization resolution.
The use of k-means clustering to optimize the split for weights and biases refines this process further.
\splitquant can be integrated with other quantization algorithms to enhance their performance.
Tests on two fine-tuned BERT-Tiny language models demonstrated significant improvements of 3.3\%p and 2.1\%p in accuracy with INT2 quantization, achieving accuracies comparable to the original FP32 models.
Future research could explore the application of \splitquant to large language models and investigate potential benefits from advancements in sparse DNN technologies.
\splitquant is open source and it can be downloaded at its online repository. \footnote{\urlstyle{tt}\url{https://github.com/jaewoosong/splitquant}}

\bibliographystyle{ACM-Reference-Format}
\bibliography{splitquant_bibfile}

\end{document}